\documentclass[11pt]{article}
\usepackage{coling2020}
\usepackage[utf8]{inputenc}

\usepackage{times}
\usepackage{url}
\usepackage{latexsym}
\usepackage{graphicx}
\usepackage{float}
\graphicspath{ {./images/} }

\usepackage{caption}
\usepackage{subcaption}

\colingfinalcopy 

\title{German's Next Language Model}

\author{
	Branden Chan$^{*\dagger}$, 
	Stefan Schweter$^{*\ddagger}$, 
	Timo Möller$^
	{\dagger}$ \\
	\\
	$^{\dagger}$deepset\\
	{\tt
		\{branden.chan, timo.moeller\}@deepset.ai}\\
	\\
	
	$^{\ddagger}$Bayerische Staatsbibliothek M{\"u}nchen\\
    Digital Library/Munich Digitization Center \\
    {\tt stefan.schweter@bsb-muenchen.de}\\
}

\date{June 2020}

\begin{document}

\maketitle

\blfootnote{
    \hspace{-0.65cm}  
    This work is licensed under a Creative Commons Attribution 4.0 International Licence. Licence details: http://creativecommons.org/licenses/by/4.0/.
}

\begin{abstract}

In this work we present the experiments which lead to the creation of our BERT and ELECTRA based German language models, GBERT and GELECTRA. By varying the input training data, model size, and the presence of Whole Word Masking (WWM) we were able to attain SoTA performance across a set of document classification and named entity recognition (NER) tasks for both models of base and large size. We adopt an evaluation driven approach in training these models and our results indicate that both adding more data and utilizing WWM improve model performance. By benchmarking against existing German models, we show that these models are the best German models to date. Our trained models will be made publicly available to the research community. 

\end{abstract}

\renewcommand{\thefootnote}{\fnsymbol{footnote}}
\footnotetext[1]{Equal contribution.}
\renewcommand{\thefootnote}{\arabic{footnote}}

\section{Introduction}

Deep transformer based language models have shown state-of-the-art results for various Natural Language Processing tasks like text classification, NER and question answering \cite{devlin-etal-2019-bert}. They are pretrained, first by feeding in large unlabeled text corpora before being fine-tuned on the downstream task. In this work we present a set of German BERT and ELECTRA models, the best of which, GELECTRA\textsubscript{Large}, significantly improves upon state of the art performance on the GermEval18 hate speech detection task by about +4\% / +2.5\% for the coarse and fine variants of the task respectively. This model also reaches SoTA on the GermEval14 NER task, outperforming the previous best by over +4\%. While performant, such models are prohibitively large for many and so we also present a new GBERT model which matches DBMDZ BERT, the previous best German BERT, in size but outperforms it by +2.18\% F1 averaged over three tasks.

In the process of pretraining the language models, we also a) quantify the effect of increasing the training data by an order of magnitude and b) verify that whole word masking has a positive effect on BERT models.

Because of the computational expense of training large language models from scratch, we adopt a downstream-oriented evaluation approach to ensure that we get the best performance from a limited number of runs. This involves regularly checkpointing the model over the course of pretraining, evaluating these on a set of classification and NER tasks and selecting as final the checkpoint which shows the best performance. This stands in contrast to approaches where the final model is simply saved after a fixed number of steps. Our method is also an important tool in diagnosing pretraining and we hope that it will be of use to other teams looking to train effective language models on a budget.

\section{Related work}
\label{sec:related-work}

Modern language model architectures are trained to build word representations that take into consideration the context around a given word. First versions such as ELMo \cite{elmo}, ULMFiT \cite{ulmfit} and \textsc{Flair} \cite{flair} are LSTM based and these were able to set new
performance benchmarks on downstream tasks like text classification, PoS tagging and NER. More recent
approaches use Transformer-based \cite{NIPS2017_7181} architectures and examples include GPT-2 \cite{radford2019language}, BERT \cite{devlin-etal-2019-bert}, RoBERTa \cite{liu2019roberta}, ALBERT \cite{Lan2020ALBERT} and ELECTRA \cite{Clark2020ELECTRA}.

In this work we focus on BERT and ELECTRA models. BERT uses a masked language modeling (MLM) strategy to corrupt
an input sentence by replacing some tokens with a {\tt [MASK]} symbol. The model is then trained to re-construct
the original token. However, this method of training is somewhat restricted in that the model only learns from the masked out tokens which typically make up about 15\% of the input tokens.

ELECTRA addresses this problem by introducing a new pretraining task called Replaced Token detection. Instead of masking out tokens,
a subset of the input tokens are substituted by a synthetically generated token. The model is then trained to classify whether each input token
is original or substituted, thus allowing for gradient updates at every input position. Practically
speaking, this is achieved by having a discriminator that performs the replaced token detection and a generator which provides plausible 
token substitutes. These two components are trained jointly and are both Transformer based. 

The BERT model received an update when the original authors added Whole Word Masking\footnote{\url{https://github.com/google-research/bert/commit/0fce551}} whereby masking one subword token requires that all other
tokens in the word are also masked out. The authors report that this method improves the training signal by removing the easiest cases and show that it improves performance in their tasks.

There is also a line of work that looks into bringing language modeling techniques that were first developed on English to other languages. These include but are not limited to monolingual models such as CamemBERT \cite{martin2020camembert} and FlauBERT \cite{FlauBERT} for French, Finnish BERT \cite{2019arXiv191207076V} and German BERTs by DBMDZ\footnote{\url{https://github.com/dbmdz/berts}} and deepset\footnote{\url{https://deepset.ai/german-bert}}. For a more comprehensive list, see \cite{2020arXiv200302912N}. 

Some models are also capable of supporting multiple languages such as multilingual BERT (mBERT\textsubscript{Base}) and XLM-RoBERTa \cite{xlmr}. Multilingual BERT is a multilingual model for 104 different languages\footnote{\url{https://github.com/google-research/bert/blob/f39e88/multilingual.md}} trained on Wikipedia dumps. The
XLM-RoBERTa model is trained on 2.5TB of data from a cleaned Common Crawl corpus \cite{wenzek-etal-2020-ccnet} for 100 different languages.



It is worth emphasizing here that systems trained on naturally occurring data will learn pre-existing cultural biases around gender \cite{NIPS2016_6228}, race and religion \cite{Speer}. Critical evaluation of machine learning methods is more important than ever as NLP is gaining broader adoption. Researchers have been advocating for better documentation of decisions made during the construction of a dataset \cite{2018arXiv180309010G}, explicit statements of a dataset’s “ingredients” \cite{2018arXiv180503677H} and recognition of the dataset characteristics that may lead to exclusion, overgeneralisation and underexposure \cite{bender-friedman-2018-data}. These topics will be addressed in Section \ref{sec:pretraining-data}.

\section{Datasets}

\subsection{Pretraining Data}
\label{sec:pretraining-data}

We have available to us, a range of different German language corpora that we use in different combinations for our model pretraining. OSCAR \cite{OrtizSuarezSagotRomary2019} is a set of monolingual corpora extracted from Common Crawl. The Common Crawl texts are pre-processed (e.g. HTML entities are removed) and a language classification model is used to sort texts by language. We use the unshuffled version of the German OSCAR corpus, resulting in 145GB of text. The Wikipedia dump for German is preprocessed with the WikiExtractor\footnote{\url{https://github.com/attardi/wikiextractor}} script forming a corpus of size 6GB. The OPUS project\footnote{\url{http://opus.nlpl.eu}} \cite{opus} has collected texts from various domains such as movie subtitles, parliament speeches and books and these comprise a collection of around 10GB. From Open Legal Data\footnote{\url{http://openlegaldata.io/research/2019/02/19/court-decision-dataset.html}} \cite{old} there is a dataset of about 2.4GB of German court decisions. Table \ref{table:dataset-sizes} shows an overview over all datasets.

As discussed in Section \ref{sec:related-work}, our pretrained language models will learn pre-existing biases from the training datasets. The main portion (89\%) of our training data, namely the OSCAR dataset, uses texts scraped from the internet, which is in some respects problematic. First off, this dataset contains a lot of explicit and indecent material. While we filtered out many of these documents through keyword matching, we cannot guarantee that this method was successful in every case. Furthermore, many websites contain unverified information and any dataset containing this kind of text can lead to a skewed model that reflects commonly found lies and misconceptions. This includes gender, racial and religious biases which are found in textual data of all registers and so we advise that anyone using our model to recognise that it will not always build true and accurate representation of real world concepts. We implore users of the model to seriously consider these issues before deploying it in a production setting, especially in situations where impartiality matter, such as journalism, and institutional decision making like job applications or insurance assessments.

\begin{table}
\centering
\begin{tabular}{l|c}
\hline
\textbf{Dataset} & \textbf{Size} \\
\hline
OSCAR & 145 \\
OPUS & 10 \\
Wikipedia & 6 \\
OpenLegalData & 2.4 \\
\hline
\end{tabular}
\caption{The size of each dataset in gigabytes.}
\label{table:dataset-sizes}
\end{table}

\subsection{Downstream Data}

\subsubsection{GermEval18}

For text classification we use GermEval18 (Coarse) and GermEval18 (Fine) which are both hate speech classification tasks \cite{WiegandSiegelRuppenhofer2018}. GermEval18 (Coarse) requires a system to classify a tweet into one of two classes: {\tt OFFENSE} if the tweet contains some form of offensive language, and {\tt OTHER} if it does not. GermEval18 (Fine) extends the coarse-grained task and contains four classes: {\tt OTHER} for non-offensive tweets as well as {\tt PROFANITY}, {\tt INSULT} and {\tt ABUSE} which are all subclasses of {\tt OFFENSE} from the coarse variant of the task.

\subsubsection{GermEval14}

For NER, we use the GermEval14 \cite{benikova2014germeval} shared task. The data is sampled from German Wikipedia and News Corpora and contains over 31,000 sentences and 590,000 tokens. The dataset is one of the largest NER datasets for German and features an advanced annotation schema that allows for nested annotations. The four main classes ({\tt PERSON}, {\tt ORGANISATION}, {\tt LOCATION} and {\tt OTHER}) each have part and derivative variants (e.g. {\tt LOCpart} or {\tt PERderiv}) resulting in 12 classes in total.

\section{Training}

\subsection{Method}

To train our German BERT and ELECTRA we use the Tensorflow training scripts from the official repositories\footnote{\url{https://github.com/google-research/bert} and \url{https://github.com/google-research/electra}}. We train models that match the size of the original BERT\textsubscript{Base}, BERT\textsubscript{Large}, ELECTRA\textsubscript{Base} and ELECTRA\textsubscript{Large}. The hyperparameters used for training can be found in Table~\ref{table:hyperparams-lm-pretraining}. The base models were trained on single Google Cloud TPUs v3 (8 cores) while large models were trained on pods of 16 TPUs v3 (128 cores). 

\begin{table}
\begin{tabular}{c|c|c|c|c}
\hline
& \textbf{GBERT\textsubscript{Base}} & \textbf{GBERT\textsubscript{Large}} & \textbf{GELECTRA\textsubscript{Base}} & \textbf{GELECTRA\textsubscript{Large}} \\
\hline
max sequence length & 512 & 512 & 512 & 512 \\
batch size & 128 & 2048 & 256 & 1024\\
warmup steps (k) & 10 & 10 & 10 & 30 \\
learning rate & 1e-04 & 1e-04 & 2e-04 & 2e-4\\
checkpoint every (k) & 100 & 100 & 76.6 & 100 \\
max train steps (k) & 4000 & 1000 & 766 & 1000 \\
layers & 12 & 24 & 12 & 24\\
hidden states & 768 & 1024 & 768 & 1024\\
attention heads & 12 & 16 & 12 & 16\\
vocab size (k) & 31 & 31 & 31 & 31 \\
train time (days) & 7 & 11 & 8 & 7 \\
\hline
\end{tabular}
\caption{Hyperparameters for language model pretraining.}
\label{table:hyperparams-lm-pretraining}
\end{table}

\subsection{Models}

In total, we trained 7 separate models with different combinations of data and model size as well as Whole Word Masking (WWM) for BERT models. The German DBMDZ BERT\textsubscript{Base}, is the same size as BERT\textsubscript{Base} and was trained using the OPUS and Wikipedia corpora. It serves as our baseline model. We train four BERT variants of it, each referred to as GBERT, each using the same cased vocabulary as DBMDZ BERT\textsubscript{Base}. These match BERT\textsubscript{Base} in size unless they have the "Large" suffix, in which case they match BERT\textsubscript{Large}:

\begin{itemize}
\item GBERT\textsubscript{Data} - trained on all available data without Whole Word Masking
\item GBERT\textsubscript{WWM} - trained on the same data as DBMDZ BERT\textsubscript{Base} but uses Whole Word Masking
\item GBERT\textsubscript{Data + WWM} - trained on all available data and uses Whole Word Masking
\item GBERT\textsubscript{Large} - trained on all available data and uses Whole Word Masking
\end{itemize}

We also trained three ELECTRA variants of DBMDZ BERT\textsubscript{Base}, each referred to as GELECTRA models, which also match the size of the original ELECTRA\textsubscript{Base} unless they have the "Large" suffix in which case they match ELECTRA\textsubscript{Large}:

\begin{itemize}
\item GELECTRA - trained on same data as DBMDZ\textsubscript{Base} BERT
\item GELECTRA\textsubscript{Data} - trained on all available data
\item GELECTRA\textsubscript{Large} - trained on all available data
\end{itemize}

The best models of each architecture and size are uploaded to the Hugging Face model hub\footnote{\url{https://huggingface.co/models}} as \emph{deepset/gbert-base}, \emph{deepset/gbert-large}, \emph{deepset/gelectra-base} and \emph{deepset/gelectra-large}.

\section{Evaluation}

In our approach, models are evaluated continuously during pretraining. Model checkpoints are saved at regular intervals and converted into PyTorch models using Hugging Face's Transformers library \cite{Wolf2019HuggingFacesTS}. Using the FARM framework\footnote{\url{https://github.com/deepset-ai/FARM}}, we evaluate the performance of each checkpoint on GermEval18 (Coarse) and GermEval18 (Fine) which are both hate speech classification tasks \cite{WiegandSiegelRuppenhofer2018}. Using Hugging Face's Transformers we also evaluate on GermEval14 \cite{benikova2014germeval} which is a NER task.

In BERT, the vector corresponding to the {\tt [CLS]} token serves as a representation of the whole input sequence, while in ELECTRA, all word vectors are combined through a feed forward layer. In both cases, this input sequence representation is passed through a single layer Neural
Network in order to perform prediction. In the NER task, each vector corresponding to the first token in a word is
passed through a single layer Neural Network and the resulting prediction is applied to the whole word.

Each checkpoint is evaluated 3 times on each document classification task since we observed significant variance across different runs. Each of these runs is performed with early stopping and a different seed each time. For NER, the model is evaluated just once without early stopping. The reported performance is the average of the single best run for GermEval18 (Coarse), GermEval18 (Fine) and GermEval14. Table~\ref{table:hyperparams-finetuning} summarizes the most important details and parameters of each task. For all experiments, we use an Nvidia V100 GPU to accelerate training. For each model, we choose the checkpoint that shows the best performance.

For comparison, we also run this evaluation pipeline on the two publicly available German BERT models (deepset German BERT\textsubscript{Base} and DBMDZ German BERT\textsubscript{Base}) as well as multilingual models such as mBERT\textsubscript{Base} and XLM-RoBERTa\textsubscript{Large}.

\begin{table}
\centering
\begin{tabular}{c|c|c|c}
\hline
& \textbf{GermEval18 (Coarse)} & \textbf{GermEval18 (Fine)} & \textbf{GermEval14} \\
\hline
Type & Classification & Classification & NER \\
Train Samples & 4509 & 4509 & 24002 \\
Dev Samples & 501 & 501 & 2200 \\
Test Samples & 3532 & 3532 & 5100 \\
Classes & 2 & 4 & 12 \\
Max Epochs & 5 & 5 & 3 \\
Max Train Steps & 705 & 705 & 4500 \\
Evaluation Every & 50 steps & 50 steps & 1500 steps \\
Learning Rate & 5e-06 & 5e-06 & 5e-05 \\
Batch Size & 32 & 32 & 16 \\
Max Seq Len & 150 & 150 & 128 \\
Metric & F1 (macro) & F1 (macro) & F1 (micro) \\
\hline
\end{tabular}
\caption{Details of the downstream tasks and hyperparameters for model finetuning for all three tasks.}
\label{table:hyperparams-finetuning}
\end{table}

\newcommand\x{0.55}


\begin{figure}[ht]
\centering
\begin{subfigure}[b]{0.49\textwidth}
  \centering
  \includegraphics[scale=\x]{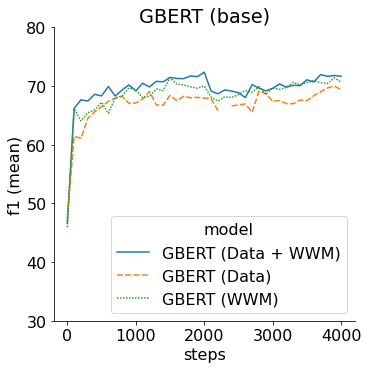} 
  \label{fig:sub-gbert-base}
\end{subfigure}
\hfill
\begin{subfigure}[b]{0.49\textwidth}
  \centering
  \includegraphics[scale=\x]{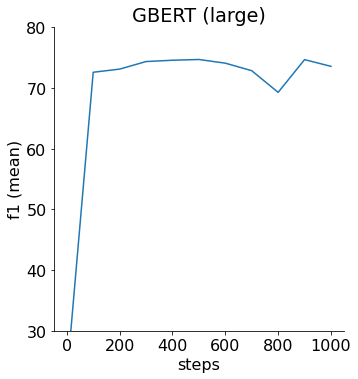} 
  \label{fig:sub-gbert-large}
\end{subfigure}
\\
\begin{subfigure}[b]{0.49\textwidth}
  \centering
  \includegraphics[scale=\x]{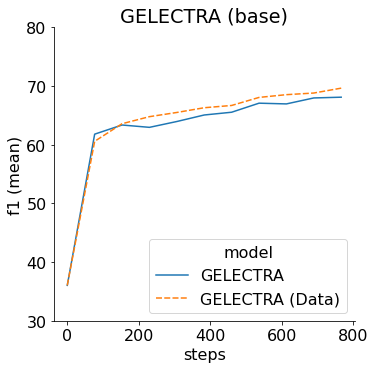} 
  \label{fig:sub-gelectra-base}
\end{subfigure}%
\begin{subfigure}[b]{0.49\textwidth}
  \centering
  \includegraphics[scale=\x]{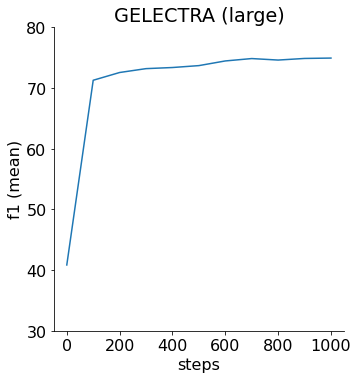} 
  \label{fig:sub-gelectra-large}
\end{subfigure}%

\caption{The F1 performance of each model averaged over the three downstream tasks over the course of language model pretraining.}
\label{figure:all-averaged-model-performance}
\end{figure}

\section{Results}

The downstream performance graphs in Figure~\ref{figure:all-averaged-model-performance} show that the models are capable of learning with most of the gains being made in the first phase of training and more incremental gains coming later. The best checkpoints come at different points for different models as can be seen in Table~\ref{table:best-checkpoint-models}.

In Table~\ref{table:downstream-evaluation-results} are the evaluation results for each model’s best checkpoint for each of the three downstream tasks with comparison to benchmark models and previous SoTA results. For GermEval18, results from the best-performing systems are reported \cite{WiegandSiegelRuppenhofer2018}. For GermEval14 we report the result that can be achieved using the \textsc{Flair} framework \cite{akbik-etal-2019-flair}.

\begin{table}[H]
\centering
\begin{tabular}{l|c}
\hline
& \textbf{Steps (k)} \\
\hline
GBERT\textsubscript{Data} & 3900 \\
GBERT\textsubscript{WWM} & 1500 \\
GBERT\textsubscript{Data + WWM} & 2000 \\
GBERT\textsubscript{Large} & 900 \\
\hline
GELECTRA & 766 \\
GELECTRA\textsubscript{Data} & 766 \\
GELECTRA\textsubscript{Large} & 1000 \\
\hline
\end{tabular}
\caption{Best checkpoint of each trained model.}
\label{table:best-checkpoint-models}
\end{table}

\begin{table}
\resizebox{\textwidth}{!}{
\begin{tabular}{l|c|c|c|c|c}
\hline
& Params & \textbf{GermEval18 (Coarse)} & \textbf{GermEval18 (Fine)} & \textbf{GermEval14} & \textbf{Averaged F1} \\
\hline
DBMDZ BERT\textsubscript{Base} & 110m & 75.23 & 47.39 & 87.90 & 70.17 \\
deepset BERT\textsubscript{Base} & 110m & 74.7 & 48.8 & 86.87 & 70.12 \\
\hline
mBERT\textsubscript{Base} & 172m & 70.00 & 45.20 & 87.44 & 67.55 \\
XLM-Roberta\textsubscript{Large} & 550m & 78.38 & 54.1 & 87.07 & 73.18 \\
\hline
GBERT\textsubscript{Data} & 110m & 74.51 & 48.01 & 87.41 & 69.97 \\
GBERT\textsubscript{WWM} & 110m &76.48 & 49.99 & 87.80 & 71.42 \\
GBERT\textsubscript{Data + WWM} & 110m & 78.17 & 50.90 & 87.98 & 72.35 \\
GBERT\textsubscript{Large} & 335m & 80.08 & 52.48 & 88.16 & 73.57 \\
\hline
GELECTRA & 110m & 76.02 & 42.22 & 86.02 & 68.09 \\
GELECTRA\textsubscript{Data} & 110m & 76.59 & 46.28 & 86.02 & 69.63 \\
GELECTRA\textsubscript{Large} & 335m & \textbf{80.70} & \textbf{55.16} & \textbf{88.95} & \textbf{74.94}\\
\hline
Previous SoTA && 76.77 (TU Wien) & 52.71 (uhhLT) & 84.65 (\textsc{Flair}) & \\
\hline
\end{tabular}
}
\caption{Downstream evaluation results for the best checkpoints of each GBERT and GELECTRA model compared to a set of benchmark models. For GermEval18 we report scores for the best-performing systems \cite{WiegandSiegelRuppenhofer2018}, and the result reported by \textsc{Flair} framework \cite{akbik-etal-2019-flair} for GermEval14.}
\label{table:downstream-evaluation-results}
\end{table}

In GermEval18 (Coarse), GBERT\textsubscript{Data + WWM}, XLM-Roberta\textsubscript{Large}, GBERT\textsubscript{Large} and GELECTRA\textsubscript{Large} all improve upon the previous SoTA. GELECTRA\textsubscript{Large} does so with the largest margin reaching a score that is +3.93\% better. In GermEval18 (Fine), XLM-Roberta\textsubscript{Large} beats the previous best by +1.39\% and GELECTRA\textsubscript{Large} sets a new SoTA that is better than the previous by +2.45\%. In GermEval14, all 7 trained models exceed the previous SoTA, with GELECTRA\textsubscript{Large} showing a +4.3\% improvement over the previous best.

These results indicate that adding extra data generally gives a modest performance boost to our language models. GBERT\textsubscript{Data + WWM} outperforms GBERT\textsubscript{WWM} by +0.93\% and GELECTRA\textsubscript{Data} outperforms GELECTRA by +1.59\%. However, GBERT\textsubscript{Data} performs worse than DBMDZ BERT\textsubscript{Base} by -0.2\%. For the BERT models, Whole Word Masking also shows a consistent positive impact with GBERT\textsubscript{WWM} outperforming DBMDZ BERT\textsubscript{Base} by +1.25\% and GBERT\textsubscript{Data + WWM} outperforming GBERT\textsubscript{Data} by +2.38\%.

\section{Discussion}

\subsection{Model Size}

The large models that we train show much stronger performance than the base models. GBERT\textsubscript{Large} outperforms GBERT\textsubscript{Data + WWM} by +2.33\% averaged F1 and GELECTRA\textsubscript{Large} outperforms GELECTRA\textsubscript{Data} by +5.31\%. It must be noted however, that their differing training regimes mean that the large models are trained on many more tokens than their base counterparts. In future, we would also be interested in training larger models with less data in order to better quantify the gains that come from model size and the gains that come from the extra data.

\subsection{Training Length}
From the downstream evaluation graphs in Figure \ref{figure:all-averaged-model-performance}, it is clear that the models gain most of their performance after a relatively short amount of training steps. GBERT\textsubscript{WWM} and GBERT\textsubscript{Data + WWM} both show an upward trend in the second half of model training suggesting they could still benefit from continuing training. There is also a clear upward trend over the course of GELECTRA and GELECTRA\textsubscript{Data}'s training suggesting these models are undertrained. It should also be noted that none of the models exhibit any clear signs of overfitting or performance degradation and may improve with further training.

\subsection{ELECTRA Efficiency}

One of the central claims of the ELECTRA paper is that it is capable of learning more efficiently than MLM based Language Models. This is exemplified by the comparison of GBERT\textsubscript{Large} and GELECTRA\textsubscript{Large}. By the end of their 1 million steps of training, GELECTRA\textsubscript{Large} has only seen half the number of tokens that GBERT\textsubscript{Large} due to its smaller batch size and yet outperforms it by +1.47\% averaged F1.

\subsection{Instabilities}
The dip in performance around 2 million steps for the base sized GBERT models (See Figure \ref{figure:all-averaged-model-performance}) happens to coincide with our training regime whereby the model training is stopped, saved and then reloaded at 2 million steps. While we suspect that these two events are related, it was beyond the scope of this project to investigate the exact reasons.



\section{Conclusion}
The set of German models which we trained vary in terms of training regime and model architecture. We hope that the results that we present here will serve as important data points to other NLP practitioners who are looking to train language models from scratch but are limited by compute. Our experiments should give other teams a sense of the batch sizes and training lengths that make for efficient model training. On top of this, we also present a set of GELECTRA and GBERT models which, according to our evaluations, set new SoTA performance for both large and base sized models on GermEval18 and GermEval14.

\section*{Acknowledgements}
We would like to thank the deepset team, especially Malte Pietsch and Tanay Soni for their regular sparring and their effort maintaining FARM. Thanks to Zak Stone, Jonathan Caton and everyone at the Google TensorFlow Research Cloud team for their advice and for providing us with the access to and credits for the TPU pods that we used for pretraining. We would also like to thank Nikhil Dinesh from the AWS Activate program as well as Nvidia's Inception program for providing us with the EC2 instances and credits that allowed us to do large scale evaluation of our models. Thanks also to Pedro Javier Ortiz Su{\'a}rez and the OSCAR corpus team for giving us access to their dataset. And thanks to Malte Ostendorff, co-founder of Open Justice e.V., whose team created Open Legal Data.


\bibliographystyle{coling.bst}
\bibliography{coling2020.bib}

\end{document}